\documentclass{article} 
\usepackage{iclr2024_conference,times}

\usepackage{hyperref}
\usepackage{url}
\usepackage{booktabs}
\usepackage{makecell}
\usepackage{multirow}
\usepackage{tabularx}

\usepackage{graphicx}
\usepackage{siunitx}
\usepackage{svg}
\usepackage{xcolor}
\newcommand*{\rom}[1]{\expandafter\@slowromancap\romannumeral #1@}

\usepackage{hyperref}
\def\dataurl{\url{ https://doi.org/10.25625/QUTUWU}}

\title{Towards general deep-learning-based tree instance segmentation models}

\author{Jonathan Henrich\\
Chairs of Statistics and
Econometrics\\
Faculty of Economics\\
University of Göttingen \\
Germany\\
\texttt{jonathan.henrich@uni-goettingen.de}
\And
Jan van Delden \\
\\
Institute of Computer Science \\  University of Göttingen \\
Germany \\
\texttt{jan.vandelden@uni-goettingen.de}
}

\iclrfinalcopy
\begin{document}

\maketitle

\begin{abstract}
The segmentation of individual trees from forest point clouds is a crucial task for downstream analyses such as carbon sequestration estimation. Recently, deep-learning-based methods have been proposed which show the potential of learning to segment trees. Since these methods are trained in a supervised way, the question arises how general models can be obtained that are applicable across a wide range of settings. So far, training has been mainly conducted with data from one specific laser scanning type and for specific types of forests. In this work, we train one segmentation model under various conditions, using seven diverse datasets found in literature, to gain insights into the generalization capabilities under domain-shift. Our results suggest that a generalization from coniferous dominated sparse point clouds to deciduous dominated high-resolution point clouds is possible. Conversely, qualitative evidence suggests that generalization from high-resolution to low-resolution point clouds is challenging. This emphasizes the need for forest point clouds with diverse data characteristics for model development. To enrich the available data basis, labeled trees from two previous works were propagated to the complete forest point cloud and are made publicly available at \dataurl. 
\end{abstract}

\section{Introduction}

As global climate change accelerates, driven by anthropogenic activities, the role of forests in carbon sequestration, biodiversity preservation, and regulation of local and global climatic conditions has been brought into sharp focus. To investigate how forests contribute to these environmental aspects, quantifiable data on their structure and development is urgently needed. In this context, technologies that enable the creation of holistic, three-dimensional representations of forests in the form of point clouds play a vital role. Such technologies are terrestrial or mobile laser scanning (TLS, MLS), but also laser scanning via low-flying unmanned aerial vehicles (UAV). Such forest point clouds often need to be segmented into individual trees for further analysis, which is an instance segmentation problem. The most commonly used paradigm for tree segmentation is to first detect tree trunks and then assign the remaining points to individual trees based on hand-crafted features such as distance or local geometry \citep{trochta20173d, burt2019extracting}. However, laser scanning characteristics, forest structures, and interactions between trees are diverse. So, defining a fixed set of assignment rules and features that consistently lead to a good segmentation performance is a highly challenging task. 

Advances in point cloud processing outside the forest domain show the advantage of performing instance segmentation using deep learning \citep{vu2022softgroup, jiang2020pointgroup}, so that relevant features can be learned in a data-driven way. Only recently, these methods have been applied to the forest domain, yielding promising segmentation results \citep{xiang2023towards, henrich2023treelearn}. Since these methods are trained in a supervised way using specific datasets, a key challenge is to obtain general models that are applicable across a wide range of settings. In this context, an important question is how models generalize to out-of-domain settings. 
Differences in the training data are, for example, caused by different laser scanning characteristics or forest types. Training a supervised deep learning algorithm requires forest point clouds that come with segmentation labels. Although recent works have acknowledged this need and put considerable effort into making high-quality labeled forest point clouds publicly available \citep{puliti2023instance, henrich2023treelearn}, the size and diversity of these datasets is still limited. Other works provide segmented trees that have been manually segmented \citep{tockner2022automatic} or manually checked for quality assurance \citep{calders2022laser}. However, these works do not include the non-tree points in their published data. Only if labels are available for the complete point cloud, it is possible to train a fully deep learning-based segmentation pipeline that does not require separate pre-processing steps.

This work makes two contributions: (1) The existing corpus of labeled forest point clouds is extended by propagating the publicly provided individual tree labels of two previous works \citep{tockner2022automatic, calders2022laser} to the complete point clouds. These point clouds are made publicly available. (2) An existing deep-learning-based tree segmentation model \citep{henrich2023treelearn} is trained with forest point clouds from different settings to provide insights into the generalization capabilities under domain-shift.

\section{Materials and Methods}

\newcolumntype{Y}{>{\centering\arraybackslash}X}
\newcolumntype{Z}{>{\scriptsize}X}

\begin{table*}[t!] 
\centering
\scriptsize
\caption{Summary of the characteristics of the forest data used in this work. Number in parantheses of column \emph{n} trees denotes number of trees of at least 10~m height. Summary for NIBIO, CULS, TU\_WIEN and SCION is taken from \cite{puliti2023instance}. \label{tab:plots}}
\begin{tabularx}{\linewidth}{lYYZYYYZZ} 
\toprule
    & Name  & Country & {Reference} & $n$ plots & $n$ trees & \makecell{Annotated\\area ($ha$)} & Forest type & Sensor \\
\midrule
    & L1W & Germany & \citet{henrich2023treelearn} & 1 & 200 (200) & 1.16 & temperate deciduous forest & ZEB-Horizon \\
    & NIBIO & Norway & \citet{puliti2023tree} & 20 & 575 (482) & 1.21 & coniferous dominated boreal forest & Riegl miniVUX-1 UAV\\
    & CULS & Czech Republic & \citet{kuvzelka2020very} & 3 & \phantom{5}47 (47) & 0.33 & coniferous dominated temperate forest & Riegl VUX-1 UAV\\
    & TU\_WIEN & Austria & \citet{wieser2017case} & 1 & 150 (106) & 0.55 & deciduous dominated alluvial forest & Riegl VUX-1 UAV\\
    & SCION & New Zealand & Unpublished & 5 & 135 (130) & 0.33 & non-native pure coniferous temperate forest & Riegl MiniVUX-1 UAV\\
    & RMIT & Australia & Unpublished & 1 & 223 (92) & 0.37 & Native dry sclerophyll eucalypt forest & Riegl MiniVUX-1 UAV\\
    \midrule
    & LAUTX & Austria & \citet{tockner2022automatic} & 6 & 514 (354) & 0.83 & temperate mixed forest & ZEB-Horizon\\
    & WYTHAM & England & \citet{calders2022laser} & 1 & 877 (608) & 1.52 & temperate deciduous forest & RIEGL VZ-400\\
\bottomrule
\end{tabularx}
\end{table*}

\subsection{Labeled forest data}\label{sec:data}

The TreeLearn method can be trained on complete labeled forest point clouds that have a sufficiently high scanning resolution for all parts of a tree. The existing literature was searched for data that fulfils this criterion. First, there is the recently published FOR-instance dataset \citep{puliti2023instance} in which tree labels and fine-grained semantic labels were manually added to point clouds from existing works. These point clouds have been captured via UAV-laser scanning and consist of diverse forest plots located in Norway (NIBIO), Czech Republic (CULS), Austria (TU\_WIEN), New Zealand (SCION) and Australia (RMIT). In another recent work, tree labels for a forest plot located in Germany (L1W) were obtained using the Lidar360 software \citep{Lidar360} and then manually corrected. A summary of the characteristics of each dataset can be found in Table~\ref{tab:plots}. More precise information can be found in the respective publications.

Apart from these point clouds, two published datasets were identified that consist of high-quality segmented trees obtained by an automatic segmentation algorithm that were either manually checked \citep[WYTHAM,][]{calders2022laser} or corrected \citep[LAUTx,][]{tockner2022automatic} for quality assurance. The respective authors were contacted to obtain the complete unlabeled point clouds. These point clouds additionally contain non-tree points, i.e. belonging to the understory or ground, and non-annotated points, i.e. points that belong to trees but have not been annotated in the published datasets. For example, some parts of the tree crown that are hard to clearly assign to a specific tree might not have been annotated.

To obtain labels for the complete point clouds, the tree labels from the published datasets have to be propagated and the remaining points must be assigned to the classes ``non-tree'' or ``non-annotated''. This was done as follows:
\begin{enumerate}
    \item For each point in the unlabeled forest point cloud, the most common tree label within a 0.1~m radius was assigned.
    \item Among the remaining unlabeled points, non-tree points were identified using proximity-based clustering: All points that were within a 0.3~m distance to each other were linked and the largest connected component was labeled as non-tree points. The large grouping radius together with the high resolution of the point clouds ensured that all understory and ground points were added to the non-tree class.
    \item The points that were still unlabeled at this stage represent tree points that have not been annotated and were assigned to the non-annotated class. This information can be used to disregard these points during training. 
    \item Finally, we visually inspected the point clouds to ensure that they were adequately divided into trees, non-tree points and non-annotated points. Remaining errors were manually corrected within a feasible scope. Specifically, one large tree was not segmented in the original labeled data of \cite{calders2022laser} which was added, and the tree bases of \cite{tockner2022automatic} were corrected since they were only roughly segmented in the original labeled data.
\end{enumerate}
For the given datasets, high-quality segmentation labels are only ensured when considering trees larger than 10~m, while assigning the rest as non-trees. In WYTHAM, smaller trees are inconsistently labeled, i.e. sometimes as a tree and sometimes as non-tree. In LAUTX, smaller trees have severe quality limitations. A correction of these mistakes was beyond the scope of this work. Therefore, only trees larger than 10~m were considered here.

\subsection{Segmentation Method}

The model framework used in this study is TreeLearn \citep{henrich2023treelearn}. It employs the widely-used grouping-based paradigm \citep{qi2019deep} for instance segmentation: The point cloud is processed using a 3D-UNet followed by pointwise semantic and offset prediction. The semantic prediction is used to classify points as tree or non-tree. The offset prediction aims to shift each point towards the respective tree base a point belongs to. After applying the predicted offset to each point, tree instances can be identified using density-based clustering. To account for memory limitations, the authors proposed a sliding window approach with subsequent merging of the results.

\subsection{Experiments}

Using the labeled data presented in Section~\ref{sec:data}, TreeLearn was trained in three conditions: (i) In the first condition, only UAV-data was used (NIBIO, CULS, TU\_WIEN, SCION). Most of these point clouds come from coniferous dominated forests. (ii) In the second condition, only TLS and MLS data (LAUTX, WYTHAM) were used, which come from mixed or deciduous forests. (iii) Lastly, all data was used for model training. In all three conditions, an area covering roughly 400 trees from WYTHAM was employed as the validation set. The number of trees in the training data in condition (i) and (ii) is roughly equal (765 vs. 762). Test performance was evaluated using L1W, a beech-dominated deciduous forest. Condition (i) assesses the effect of using out-of-domain data during training since the laser scanning characteristics and tree composition are substantially different from L1W. Condition (ii) represents in-domain data. In addition to quantitative test results on L1W, qualitative test results on a low-resolution UAV point cloud (RMIT) are presented.

In each condition, the initial model weights were set to a publicly available TreeLearn checkpoint that has been obtained by training on large amounts of MLS data with non-corrected labels from a commercial software. From there, fine-tuning was performed for \num[group-digits=integer]{12500} iterations using the AdamW optimizer \citep{adamw} with a weight decay of $10^{-3}$ and $\beta = [0.9, 0.999]$. The batch size was set to 2. A cosine learning rate schedule \citep{cosinelr} with a maximum/minimum learning rate of $1 \times 10^{-3}$/$5 \times 10^{-5}$ was selected. Training examples were generated by randomly cropping squares of size 35~m by 35~m from the labeled forest point clouds. Only the inner 8~m by 8~m were considered during gradient computation so that the respective tree base of a tree point was ensured to be within the crop.

The performance on the L1W-dataset is evaluated based on the evaluation protocol detailed in \cite{henrich2023treelearn}. First, the tree detection performance is measured by the number of false positive and false negative predictions. To assess the semantic segmentation into tree and non-tree points, the accuracy is calculated. Instance segmentation performance is evaluated using the F1-score. It is calculated for each tree separately based on the number of true positive, false positive and false negative points and then averaged across all trees.

\section{Results and discussion} \label{sec:results}

\begin{table*}[t!] \label{table:res}
\begin{minipage}{\textwidth}
\center
\caption{Segmentation results on L1W. Semantic and instance segmentation results in~\%. Results of TreeLearn trained on Lidar360 labels are taken from \cite{henrich2023treelearn}. \label{tab:testresults}}
\begin{tabular}{lccccc}
\toprule
& \multicolumn{1}{c}{Semantic Seg.} & \multicolumn{2}{c}{Detection} & Instance Seg.\\
\cmidrule(lr){2-2}
\cmidrule(lr){3-4}
\cmidrule(lr){5-5}
	Training Data & Accuracy  & $FP$ Predictions & $FN$ Trees & $F1$ \\
\midrule
MLS (Lidar360 labels)                           & \textbf{99.69} & 0 & 0 & 93.98 \\
\midrule
+UAV           & 99.41 & 1 & 0 & 96.25 \\
+MLS+TLS           & 99.64 & 1 & 0 & \textbf{97.31} \\
+UAV+MLS+TLS           & 99.64 & 0 & 0 & 96.88 \\
\bottomrule
\vspace{0pt}

\end{tabular}
\end{minipage}
\end{table*}

\begin{figure*}[h]
   \centering
    \includegraphics[width=\textwidth]{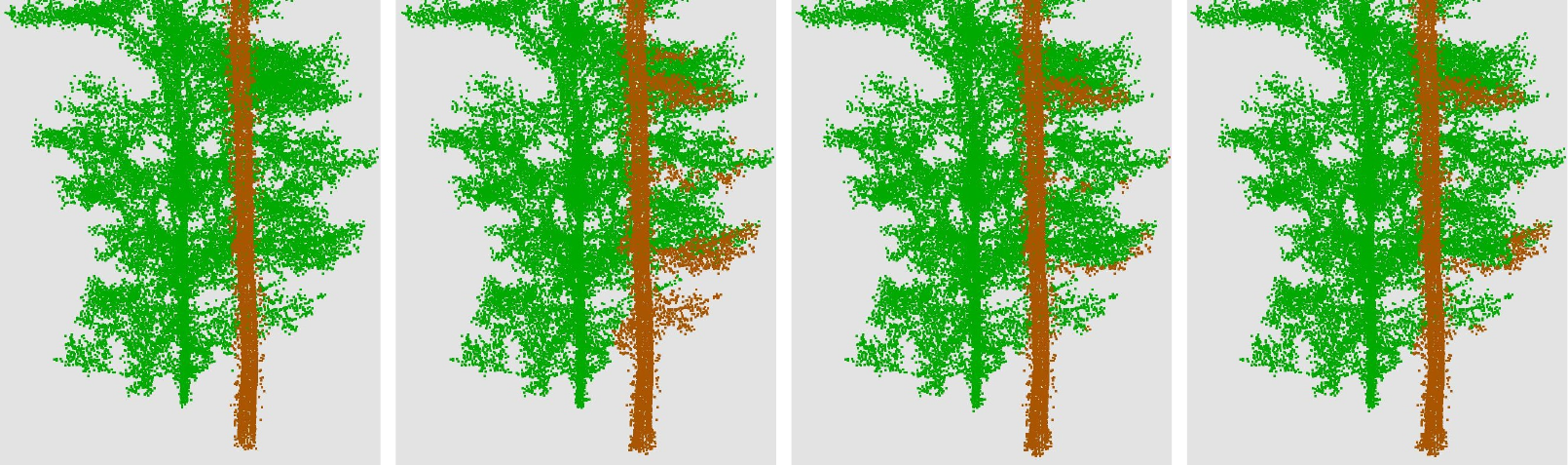}
    \caption[]{Fine-grained test results on an MLS point cloud (L1W). From left to right the images show (1) the ground truth segmentation and model results obtained by fine-tuning on (2) UAV, (3) MLS+TLS and (4) all data. Results are best when in-domain data is included during training.}
    \label{fig:instance_fig}
\end{figure*}

\begin{figure*}[h]
   \centering
    \includegraphics[width=\textwidth]{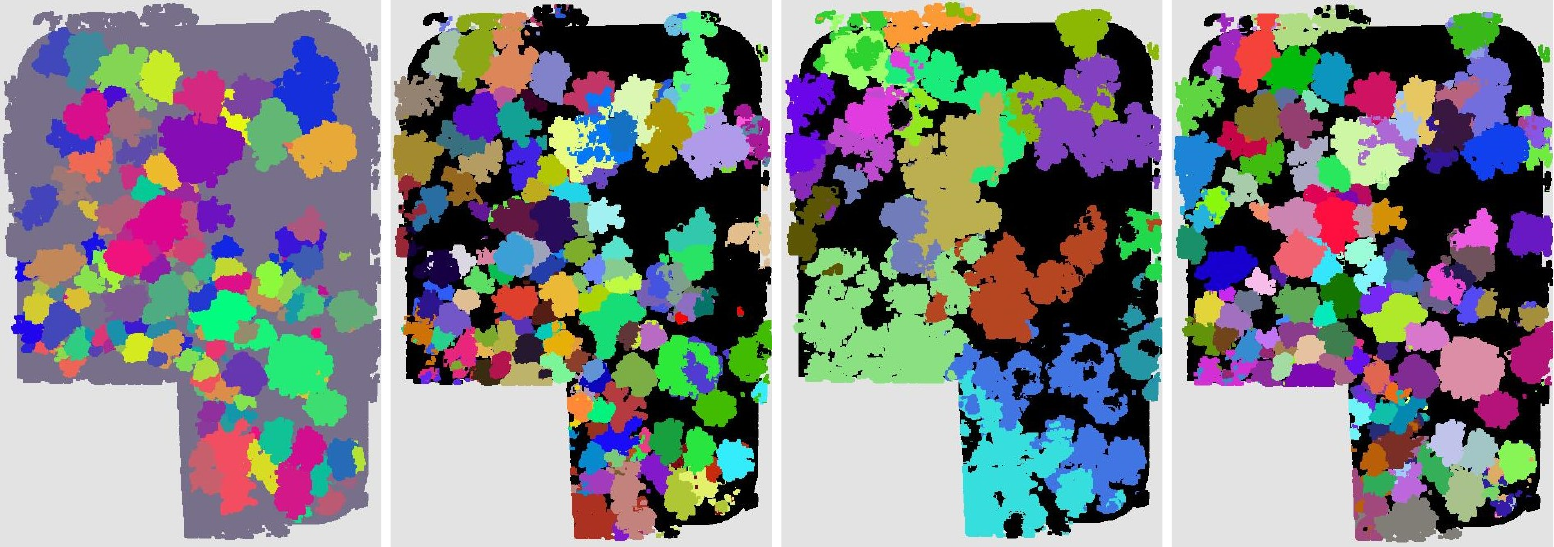}
    \caption[]{Qualitative test results on a low-resolution UAV-scanned point cloud (RMIT). From left to right the images show (1) the ground truth segmentation and model results obtained by fine-tuning on (2) UAV, (3) MLS+TLS and (4) all data. When only MLS and TLS data is used, segmentation results have severe mistakes. When UAV data is included during training, results are substantially improved.}
    \label{fig:rmit_fig}
\end{figure*}

Segmentation results were computed for the point cloud L1W (MLS, deciduous dominated). Results obtained by using the publicly available TreeLearn checkpoint without further training serve as a baseline (Table~\ref{table:res}). Even when fine-tuning the model with out-of-domain data (UAV, coniferous dominated), instance segmentation performance in terms of the F1-score increases substantially from 93.98~\% to 96.25~\%. In terms of tree detection (one FP) and semantic segmentation (99.41\%), the fine-tuned model performs slightly worse than the baseline. However, these aspects are less important compared to instance segmentation performance. For example, FP predictions can be manually discarded or merged into complete trees without much effort if they are not too frequent. When using in-domain data (MLS+TLS, deciduous dominated) for fine-tuning, instance segmentation performance is further increased to 97.31\% (see Fig. \ref{fig:instance_fig} for a qualitative comparison). Since the number of trees used during training is roughly equal in the in-domain and out-of-domain conditions, the performance gap is most likely due to the domain-shift. However, other factors, such as differences in forest complexity, are hard to quantify and cannot be controlled for. When using all available data for training, test performance decreases slightly (96.88~\%) compared to only using in-domain data. In addition to quantitative segmentation results on L1W, all three models were used to obtain qualitative results on RMIT, a low-resolution UAV point cloud (Fig. \ref{fig:rmit_fig}). These results suggest that an adequate segmentation performance on low-resolution UAV data can only be achieved when including it during training. If only MLS+TLS data is used, segmentation quality decreases drastically due to severe cases of merged trees.

\section{Conclusion}

In this paper, we trained the deep-learning-based tree segmentation method TreeLearn with data from various domains and systematically evaluated its test performance. It was shown that a model trained on out-of-domain coniferous dominated UAV point clouds can generalize to deciduous dominated MLS point clouds. 
Qualitative results indicate that training exclusively with high-resolution data, although improving performance in this domain, leads to poor generalization in low-resolution UAV settings (Fig. \ref{fig:rmit_fig}).
Including UAV data in addition to high-resolution data during training alleviates this issue. This emphasizes the importance of a broad training data basis to obtain models that are applicable to a wide range of domains. To enrich the available forest point clouds, labeled tree data from previous works was propagated to the whole forest point cloud and is made publicly available at \dataurl. Going forward, a quantifiable characterization of different forest point clouds should be established to enable a more thorough and systematic comparison between domains. Furthermore, the consequences of stronger domain-shifts in terms of forest structure on model performance should be investigated, for example by using dense tropical forests. Such experiments are crucial to determine what exactly is needed in terms of model development and data provision to obtain powerful and general tree segmentation models. 

The results of this study can be regarded as preliminary evidence for the potential of deep learning to obtain general tree segmentation methods. While such methods rely on high-quality labeled forest data, many recent works have acknowledged this need by providing publicly available datasets. Due to the rapid development of deep learning methods and the availability of more and more high-quality labeled data, we expect deep-learning-based tree segmentation to become an increasingly powerful tool. In contrast to traditional segmentation methods, such methods are able to learn segmentation rules for forest point clouds with diverse characteristics in a data-driven way, thus eliminating the need for cumbersome hyperparameter tuning or models designed for specific domains. These features make methods user-friendly and practically applicable given highly diverse forest and point cloud characteristics.
\newpage
\bibliography{main}

\end{document}